\documentclass{article}

\usepackage{PRIMEarxiv}

\usepackage[utf8]{inputenc} 
\usepackage[T1]{fontenc}    
\usepackage{hyperref}       
\usepackage{url}            
\usepackage{booktabs}       
\usepackage{amsfonts}       
\usepackage{nicefrac}       
\usepackage{microtype}      
\usepackage{lipsum}
\usepackage{fancyhdr}       
\usepackage{graphicx}       
\usepackage{amsmath}        
\usepackage{subcaption} 
\usepackage{graphicx} 

\graphicspath{{media/}}     

\title{NeRF-Supervised Feature Point Detection and Description
}

\author{
  Ali Youssef \\
  Department of Computer Science \\
  University College London \\
   \And
  Francisco Vasconcelos\thanks{Hawkes Institute, University College London.
} \\
  Department of Computer Science \\
  University College London \\
}

\begin{document}
\maketitle

\begin{abstract}
 Feature point detection and description is the backbone for various computer vision applications, such as Structure-from-Motion, visual SLAM, and visual place recognition. While learning-based methods have surpassed traditional handcrafted techniques, their training often relies on simplistic homography-based simulations of multi-view perspectives, limiting model generalisability. This paper presents a novel approach leveraging Neural Radiance Fields (NeRFs) to generate a diverse and realistic dataset consisting of indoor and outdoor scenes. Our proposed methodology adapts state-of-the-art feature detectors and descriptors for training on multi-view NeRF-synthesised data, with supervision achieved through perspective projective geometry. Experiments demonstrate that the proposed methodology achieves competitive or superior performance on standard benchmarks for relative pose estimation, point cloud registration, and homography estimation while requiring significantly less training data and time compared to existing approaches.

\end{abstract}


\section{Introduction}
\label{sec:intro}

Feature point detection and description under different scene viewpoints is a common starting point for many multi-view problems including Structure-from-Motion \cite{sfm}, visual SLAM \cite{vslam_survey_1,vslam_survey_2}, or visual place recognition \cite{visual_place_rec,visual_place_rec_DL}. In the past decade, different learning-based approaches to this problem \cite{Superpoint, R2D2, ALIKE, GCNv2} have replaced handcrafted techniques in many applications \cite{real-slam,lift-slam,dense-sfm}. Crucially, most of these models can be fine-tuned in a self-supervised manner on any single-view dataset. This is achieved by applying different homography warpings to the training data, simulating different viewpoints of the same scene with known point-to-point "ground-truth" mappings. While this training scheme is simple and flexible, the generated homography warpings are a crude simplification of multi-view perspectives, which can lead to limited model generalisability. 

\begin{figure}[tb]
  \centering
  \begin{subfigure}{0.3\linewidth}
    \includegraphics[height=3.5cm]{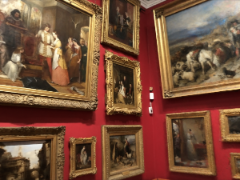}
    \caption{Input Image}
    \label{fig:short-a}
  \end{subfigure}
  \begin{subfigure}{0.3\linewidth}
    \includegraphics[height=3.5cm]{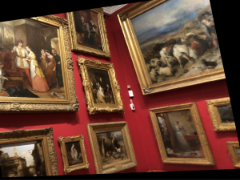}
    \caption{Image w/ Homography}
    \label{fig:short-b}
  \end{subfigure}
  \begin{subfigure}{0.3\linewidth}
    \includegraphics[height=3.5cm]{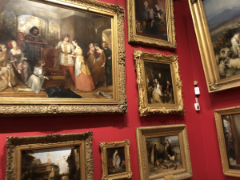}
    \caption{Image Diff. Viewpoint}
    \label{fig:short-c}
  \end{subfigure}
  \caption{\textbf{Visual representation of multi-view data.} Learning-based detectors and descriptors achieve supervision on single-view datasets by simulating different viewpoints through homographic warpings to the input image $I$ (\textit{cf.} Fig. \ref{fig:short-a}) resulting in $I^{'}$ (\textit{cf.} Fig. \ref{fig:short-b}). However, we achieve supervision by directly sampling a NeRF-rendered image from a different viewpoint (\textit{cf.} Fig. \ref{fig:short-c}).}
  \label{fig:Homography_vs_NeRF}
\end{figure}

This paper aims at leveraging image synthesis with neural radiance fields (NeRFs) as a more realistic way of generating multi-view training data for feature detection and description models (\textit{cf.} Fig. \ref{fig:Homography_vs_NeRF}). Since NeRFs require multi-view data to synthesise novel views, we can no longer rely on the single image datasets typically used for training the above-mentioned homography-based methods. Therefore, we create our dataset consisting of indoor and outdoor image sequences around static scenes and reconstruct all of them with NeRFacto \cite{NeRFStudio}. This enables the generation of arbitrary viewpoints of each scene consistent with a pinhole projective model with known point-to-point mappings via point re-projection. We propose a general methodology to upgrade state-of-the-art homography-based methods to train them on projective views synthesised from NeRF-type algorithms. Our contributions are as follows:
\begin{itemize}
    \item We create a new multi-view dataset consisting of images from 10 different indoor and outdoor scenes, and a total of 10,000 NeRF-synthesised views from these scenes with corresponding depth maps, and intrinsic and extrinsic parameters.
    \item We propose two general methodologies (end-to-end and projective adaptation) to train state-of-the-art point detection and description methods using a loss function based on NeRF re-projection error. 
    \item We re-train adapted versions of SuperPoint \cite{Superpoint} and SiLK \cite{SiLK} using our NeRF-synthesised data and compare them against the original baselines trained on the much larger MS-COCO dataset. We outperform the original baselines for relative pose estimation on ScanNet, YFCC100M and MegaDepth datasets \cite{Scannet, YFCC, Megadepth}, with similar performance for pair-wise point cloud registration, while only slightly underperforming on the HPatches homography estimation benchmark. Our code is available at \href{https://github.com/AliYoussef97/SiLK-PrP}{SiLK-PrP} and \href{https://github.com/AliYoussef97/SuperPoint-PrP}{SuperPoint-PrP}.
\end{itemize}

\section{Related Work}
\label{sec:related-work}

Research on multi-view feature detection and description has a large focus on creating representations invariant to geometric transformations (\textit{e.g.} scale, rotation, affinity) and illumination conditions so that the same scene point can be reliably recognised and matched regardless of its viewpoint. Classic methods achieved this by handcrafting feature extractors that incorporate these invariance properties by design. Some of these decade-old methods stood the test of time and are still widely used today \cite{SIFT, ORB}. More recently, the success of deep feature extraction enabled neural networks to learn these invariance properties from training data. Large open-source datasets \cite{COCO-dataset, Scannet, Megadepth} have been extensively employed to train either learning-based interest point detectors and descriptors \cite{Superpoint, DISK, SiLK, D2-Net}, or detector-free local feature matchers \cite{SuperGlue, LoFTR, COTR}. 

Some of these methods \cite{Superpoint, SiLK} are trained in a self-supervised manner on uncalibrated RGB single-view datasets such as MS-COCO \cite{COCO-dataset} by simulating multiple views as homography warpings. Other methods \cite{DISK, D2-Net} are trained with additional supervision using calibrated multi-view RGB-D datasets such as ScanNet \cite{Scannet}. These two classes of methods have pros and cons. The multi-view data generated by self-supervised methods is not representative of all possible viewpoint changes (homographies only model planar scenes or pure rotations). Thus it does not fully generalise to some application scenarios. On the positive side, they can be trained on an arbitrarily large number of transformations, and are also extremely flexible as they can be fine-tuned on any uncalibrated single-view dataset. Fully supervised methods are trained on more realistic multi-view data that covers full projective view changes with occlusions, however, the required training data is significantly more complex to obtain. The variety of scenes and viewpoint changes in the training data are also limited by the practicalities of real data acquisition. To get the best of both worlds, we propose using NeRF reconstructions obtained from RGB image sequences, which are both simple to acquire and enable the synthetic generation of arbitrary projective viewpoints beyond homographies.

Progress on image synthesis with NeRF and its variants \cite{instanerf, mipnerf, zip-nerf} has exploded in the past years, with very fast improvements in computational efficiency, image synthesis quality, reconstruction accuracy, and input data requirements. As these techniques enable synthesising arbitrary viewpoints of a reconstructed scene, they can be used for rich data augmentation \cite{feldmann2024nerfmentation}. Most related to our work, they have been utilised to self-supervise multi-view methods for learning viewpoint invariant object descriptors \cite{NeRF-descriptors}, stereo disparity \cite{NeRF-Stereo}, and optical flow \cite{AdFactory}. However, to the best of our knowledge, NeRF self-supervision has been under-explored for point feature detection and description.

\section{Methodology}
\label{sec:methodology}

\subsection{NeRF Dataset}
\label{sec:dataset}

We build a NeRF dataset for 10 indoor and outdoor scenes, containing 10,000 synthetic images with their corresponding intrinsic/extrinsic parameters and depth maps.  We note that this dataset is extremely small compared to what is typically used to train state-of-the-art feature detection  (30 times smaller than MS-COCO,~250 times smaller than ScanNet). Our goal is to show that leveraging NeRF supervision enables state-of-the-art performance on multiple benchmarks with significantly less training data.

We captured four indoor and one outdoor scene with an iPhone 10 at 4K resolution, and one indoor and one outdoor scene with a Samsung Galaxy A52 at 4k resolution. We also utilise two blender-generated indoor scenes with images sourced from \cite{blender-scenes} and one outdoor scene publicly available through \cite{mipnerf360} as they are open-source scenes that consist of high-quality RGB images. While different training sets could be constructed using existing publicly available data, we focus on a trade-off between scene diversity and minimal dataset size.

We obtain camera poses for all images using COLMAP \cite{COLMAP} and use them along with the images to train a NeRF model for synthesising synthetic views. We trained a Nerfacto model for each scene, which is part of the NeRFStudio framework \cite{NeRFStudio} and merges improvements from a range of NeRF implementations \cite{NeRF, mipnerf360, instantngp, nerf--}. For each scene, we rendered 1000 synthetic views and corresponding depth maps with $44^{\circ}$ field of view and $640\times480$ resolution. We use NeRFStudio's interactive real-time viewer to define custom camera trajectories for each scene that ensure diversity in viewpoints while maintaining global scene overlap. Trajectories consist of uniformly sampled smooth trajectories around the scene coupled with small rotations around the camera centre.

\subsection{NeRF Point Re-Projection}\label{Prp-Process}

Similarly to prior self-supervised methods \cite{Superpoint, SiLK} we consider neural networks that, at inference time, detect interest point locations $\mathbf{p}$ and extract their descriptors $\mathbf{d}$ from a single view. Consider two views $I$, $I'$ synthesised by NeRF with known intrinsic ($\mathcal{K},\mathcal{K}'$) and extrinsic ($R,R',\mathbf{t},\mathbf{t}'$) parameters. For any arbitrary point with homogeneous pixel coordinates $\mathbf{p}$ in image $I$, we know its NeRF reconstructed depth $d$ and can represent it in 3D world coordinates as: 
\begin{align}
    \mathbf{P} &= [R | \mathbf{t}] \frac{\mathbf{p}_{c}}{||\mathbf{p}_{c}||}d, \quad \mathbf{p}_{c}= \mathcal{K}^{-1}\mathbf{p}. 
    \label{eq.1} 
\end{align}

\noindent Similarly, $\mathbf{P}$ can be re-projected into view $I'$ as:
\begin{align}
    \mathbf{p'} = \mathcal{K'} \frac{\mathbf{P}{^{'}}}{\mathbf{P}{^{'}_{z}}}, \quad \mathbf{P}{^{'}} = [R'^{T} | -R'^{T}\mathbf{t'}] \mathbf{P}. 
    \label{eq.2}
\end{align}

\begin{figure}[tb]
  \centering
  
  \begin{subfigure}{0.3\linewidth}
    \includegraphics[height=4.5cm]{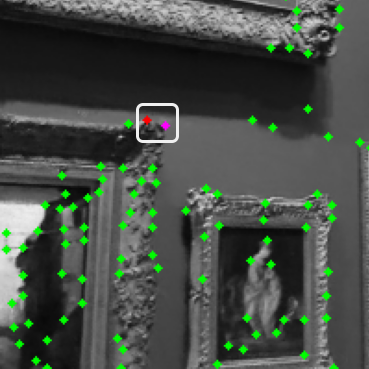}
    \caption{Input Image}
    \label{fig:depth-a}
  \end{subfigure}
  \begin{subfigure}{0.3\linewidth}
    \includegraphics[height=4.5cm]{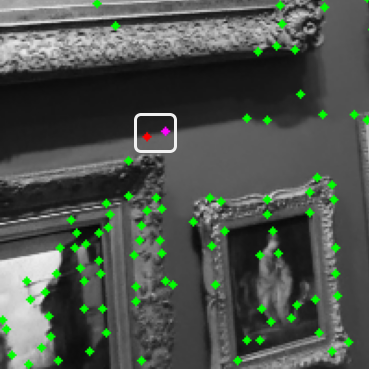}
    \caption{PrP w/o Depth Window}
    \label{fig:depth-b}
  \end{subfigure}
  \begin{subfigure}{0.3\linewidth}
    \includegraphics[height=4.5cm]{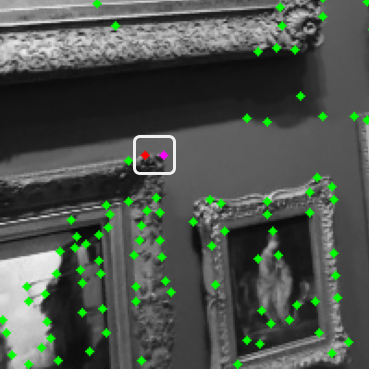}
    \caption{PrP w/ Depth Window}
    \label{fig:depth-c}
  \end{subfigure}
  
  \caption{\textbf{Depth window estimation.} The interest points, depicted in red and purple situated at the painting's frame edge in Fig. \ref{fig:depth-a}, exhibit misprojection onto image $I^{'}$ without the depth window estimation method. However, misprojection errors are effectively mitigated by utilising the depth window estimation method as seen in Fig. \ref{fig:depth-c}.}
  \label{fig:depth_rec}
\end{figure}

Assuming that reconstructed depths $d$ are accurate, the above equations enable supervising a point detector, by designing a loss function that promotes a set of re-projected points $\mathbf{p}'$ detected in view $I$ to align with points independently detected on view $I'$. However, many distinctive interest points lie on scene edges at a depth discontinuity. This means that even a small deviation in $\mathbf{p}$ can cause significant variations in $d$, corresponding to the depths of the background and foreground around the edges. To ensure stable re-projection around edges, we differentiate between foreground and background by focusing only on the foreground around the edges, excluding the background. To achieve this, we compute $d$ within a $5\times5$ pixel window centred around $\mathbf{p}$. If the difference between the maximum and minimum depths within this patch does not exceed a predefined threshold $\epsilon_d$, we take $d$ at the interest point position. Otherwise, we use the minimum depth within the window (\textit{cf.} Fig. \ref{fig:depth_rec}).

\section{Implementations}
\label{sec:Implementations}

\subsection{SiLK-PrP}\label{SiLK-prp-implementation}

Self-supervised point detection methods that are trained end-to-end can be easily adapted to utilise NeRF training data. Using SiLK \cite{SiLK} as a baseline, we propose an adapted version that, during training, simply replaces homography warpings with NeRF renderings and point re-projection, while keeping the rest of the training pipeline and loss functions intact. 

In the original SiLK approach, image pairs with ground-truth bi-directional correspondences are generated on the fly during training with homography warpings. We rely on our pre-rendered NeRF dataset described in Sec. \ref{sec:dataset}, to avoid significant computational overhead caused by on-the-fly image rendering. During training, we randomly sample image pairs $I, I^{'}$ from the same scene, Eqs. (\ref{eq.1}) and (\ref{eq.2}) are then utilised during training to generate dense ground-truth point correspondences, discretised at the pixel level.

\subsection{SuperPoint-PrP}

SuperPoint is a self-supervised method that goes through multiple stages of training. Firstly, an encoder with a detector head is trained in a fully supervised manner to detect corners on a synthetic dataset with simple geometric shapes. The pre-trained detector, called MagicPoint, is used to generate pseudo-ground truth interest points on a real dataset (\textit{e.g.} MS-COCO) using a process called homographic adaptation. This involves generating several homography warped copies of a training image, passing them through the trained detector, unwarping all predictions back to the original training image, and finally aggregating all unwarped predictions to generate the pseudo labels. This process can be repeated several times, where after each round of homographic adaptation, the detector head is further trained, updating the pseudo-ground truth labels, while improving the detector head's generalisability. After a final round of homographic adaptation, a descriptor head is added to the model, which is jointly trained with the whole model with an additional descriptor loss term on the real dataset resulting in the SuperPoint model.

With the NeRF training data, we retain the original SuperPoint architecture but adapt the homographic adaptation process and descriptor loss to handle our NeRF training data and the Point Re-Projection process.

\subsubsection{Projective Adaptation} 

We propose a novel method for generating pseudo-ground-truth labels using the PrP process instead of homography formulations. Our approach takes advantage of the continuous trajectory used to render the NeRF images.

The process begins by randomly sampling a window of 20 consecutive rendered NeRF images, denoted as $I_{i\rightarrow i+19}$. We designate the initial image $I_{i}$ as our reference and randomly select 14 of the remaining 19 images as warped versions of $I_{i}$. We refer to any image from this randomly sampled set as $I_{r}$. For both $I_{i}$ and each $I_{r}$, we compute probability heatmaps $\mathbf{H}_{i}$ and $\mathbf{H}_{r}$ respectively. We then extract interest points $\mathbf{{p}_{r}}$ from each $\mathbf{H}_{r}$ using non-maximum suppression to ensure sufficient spacing between feature points.

Next, we apply the PrP process to project the interest points $\mathbf{p}_{r}$ from each image $I_{r}$ onto $I_{i}$, resulting in projected points $\mathbf{p}_{i'}$. For all points in $\mathbf{p}_{i'}$, we create a mask $\mathbf{H}_{i'}$ by extracting a 3x3 patch centred on the original (unprojected) point location from $\mathbf{H}_{r}$ and applying it to $\mathbf{H}_{i'}$ at the projected point location. Therefore, $\mathbf{H}_{i'}$ effectively represents the probability heatmap of $I_{r}$ projected onto $I_{i}$.

Finally, we aggregate $\mathbf{H}_{i}$ with all $\mathbf{H}_{i'}$ masks and apply non-maximum suppression to the aggregated heatmaps to obtain the final set of pseudo-ground-truth interest points for $I_{i}$. This approach leverages the continuous nature of NeRF-rendered sequences to create pseudo-ground-truth labels without relying on homography approximations.

\subsubsection{Descriptor Loss}
SuperPoint's \cite{Superpoint} loss consists of an interest point detection loss $\mathcal{L}_p$ and a descriptor loss $\mathcal{L}_d$.

The interest point detection loss is a cross-entropy loss between the interest point detector prediction  $\mathcal{X}$ on the input image, and the ground truth interest points $\mathcal{Y}$. We effectively do not apply any modifications to the interest point detection loss.

SuperPoint's \cite{Superpoint} descriptor loss $\mathcal{L}_d$, is a hinge loss between every pair of descriptors cells, $d_{hw} \in \mathcal{D}$ from $\mathcal{I}$ and $d_{{h'}{w'}} \in \mathcal{D}^{'}$ from $\mathcal{I}^{'}$. "Cells" are essentially pixels in the lower dimension feature map, for an input image of size $H \times W$, the feature map has a dimension of $H_c = \frac{H}{8}$ $\times$ $W=\frac{W}{8}$. In simpler terms, each pixel in the encoded image maps an $8 \times 8$ region of the input image. SuperPoint's descriptor loss $\mathcal{L}_d$ can be defined as:

\begin{align}
     \mathcal{L}_d(D,D',S) = \frac{1}{(H_c W_c)^2} \sum^{H_c,W_c}_{\substack{h=1 \\ w=1}} \sum^{H_c,W_c}_{\substack{h'=1 \\ w'=1}} l_d(d_{hw},d'_{{h'}{w'}}; s_{hw{h'}{w'}}) \label{eq.3}
\end{align}

\noindent where $l_d$ is defined as: 

\begin{align}
    l_d(d,d';s) = \lambda_d * s * max(0,m_p - d^T d') + (1-s)*max(0,d^T d'-m_n)\label{eq.4}
\end{align}

\noindent where $m_p$ is the hinge loss's positive margin, $m_n$ is the negative margin and $\lambda_d$ balances between the positive and negative correspondences. Additionally, $S$ in Eq. (\ref{eq.3}) is a homography-induced correspondence, defined as:

\begin{align}
    s_{hw{h'}{w'}} =\left\{
    \begin{array}{ll}
      1 , & \mbox{if } ||\widehat{\mathcal{H}p_{hw}}-p_{{h'}{w'}} || \leq \epsilon_s.\\
      0, & \mbox{otherwise}
    \end{array}
    \right. \label{eq.5}
\end{align}

\noindent where $p_{hw}$ is the centre pixel in the $(h,w)$ cell, and $\widehat{\mathcal{H}p_{hw}}$ is multiplying the centre pixel cell $p_{hw}$ by a homography $\mathcal{H}$ which relates the input image and its warped pair. 

We reformulate the homography-induced correspondence of SuperPoint \cite{Superpoint}, in such a way that centre pixel $p_{hw}$ can be transformed using the PrP process rather than transforming it through homography $\mathcal{H}$. Let $\mathcal{C}(\cdot)$ represent a function that takes  $p_{hw}$ as an input, along with all parameters utilised in the PrP process (\textit{cf.} Eq. (\ref{eq.1}) and Eq. (\ref{eq.2})), for simplicity, these parameters can be combined and denoted as $M$. Therefore, the PrP-induced correspondence can be defined as:

\begin{align}
    s_{hw{h'}{w'}} =\left\{
    \begin{array}{ll}
      1 , & \mbox{if } ||\mathcal{C}(M;p_{hw})-p_{{h'}{w'}} || \leq \epsilon_s.\\
      0, & \mbox{otherwise}
    \end{array}
    \right.\label{eq.6}
\end{align}

\subsection{Implementation details}\label{Implementation-details}

For all experiments, we present results for SuperPoint-PrP and SiLK-PrP using our training pipeline, both trained exclusively on the NeRF dataset. Additionally, we report the baseline SuperPoint results based on our implementation, which closely follows the public version's training hyperparameters \cite{GlueFactory}.

To mitigate misprojection errors during the PrP process, we randomly sample image $I^{'}$ from a subset of the training data. This subset is constrained by a lower limit of $\lambda_l = 70$ and an upper limit of $\lambda_u = 150$ relative to the input image $I$, ensuring a balance between viewpoint changes while avoiding projection errors. The depth window threshold $\epsilon_d$ described in Sec. \ref{Prp-Process} is set to $3e^{-2}$ meters.

During SuperPoint-PrP and SiLK-PrP training, we apply the same photometric augmentations as baseline models \cite{Superpoint, SiLK}, but omit homographic augmentations, such as rotation, scaling or translation. All models are trained using PyTorch \cite{Pytorch} on a single NVIDIA RTX 3090 Ti GPU.

\subsubsection{SiLK-PrP training.}
SiLK-PrP is trained on the NeRF dataset in an end-to-end manner for a total of 100,000 iterations. We keep the same SiLK hyperparameters when training SiLK-PrP with ADAM optimiser, a learning rate of $1e^{-4}$ and betas as (0.9, 0.999), while the batch size was set as 1.

\subsubsection{SuperPoint-PrP training.}
Similar to the baseline training pipeline, we train MagicPoint (SuperPoint without descriptor head) on Synthetic Shapes (2D geometric shapes like lines, triangles, and cubes) for 200,000 iterations. We obtain MagicPoint-PrP, by training the MagicPoint model on the NeRF dataset for two rounds of Projective Adaptation, each for 30,000 iterations. The SuperPoint-PrP model is then obtained by training both detector and descriptor heads of MagicPoint-PrP on the NeRF for 300,000 iterations.

We use baseline SuperPoint hyperparameters with ADAM optimiser, a learning rate of $1e^{-3}$, and betas set to (0.9, 0.999). Batch sizes are 32 for MagicPoint-PrP and 2 for SuperPoint-PrP. For the baseline SuperPoint model, the default homography-induced correspondence threshold $\epsilon_s$ in Eq. (\ref{eq.5}) is 8. However, for SuperPoint-PrP, experiments showed that the optimum PrP-induced correspondence threshold $\epsilon_s$ in Eq. (\ref{eq.6}) is 4.

\section{Experiments}
\label{sec:experiments}
This section assesses the PrP-supervised models against their baselines, evaluating detection quality and homography estimation on HPatches \cite{Hpatches}, indoor relative pose estimation and point cloud registration on ScanNet \cite{Scannet}, and outdoor relative pose estimation on YFCC100M and MegaDepth \cite{YFCC, Megadepth}. All evaluations use mutual nearest neighbour matching. Through these experiments, we analyse:

\begin{itemize}
    
    \item The NeRF dataset is about $\frac{1}{30}$ the size of MS-COCO \cite{COCO-dataset} and $\frac{1}{250}$ of ScanNet \cite{Scannet}. Therefore, we evaluate the generalisability of NeRF-trained models on real-world data across various datasets. Surprisingly, the PrP models trained on the NeRF dataset perform comparably in most evaluations, questioning the need for large datasets to train learning-based detectors and descriptors.

    \item Unlike \cite{COCO-dataset, Scannet, Megadepth, Aachen}, our NeRF dataset consists entirely of synthetic images. We assess the interest point detection quality of the NeRF-trained detectors against their baselines. Experiments revealed that the quality of interest points generated by NeRF-trained models remained consistently high, showing no degradation compared to the baseline models.

    \item By rendering each NeRF scene from various viewpoints while simulating real camera motion, we evaluate whether PrP-supervised models demonstrate improvement in relative pose estimation, and require fewer training iterations than homography-based baselines. Results show that the PrP-supervised models not only exhibit enhancement in relative pose estimation but also converge faster (\textit{e.g.} SiLK-PrP in 27K iterations vs. 50K for SiLK).
    
\end{itemize}

\subsection{Homography Estimation}\label{Homography_est}
Following \cite{LoFTR, Superpoint, SuperGlue, SiLK, R2D2} we evaluate homography estimation on the HPatches dataset \cite{Hpatches}. The HPatches dataset consists of 57 scenes with varying illumination and 59 scenes with large viewpoint variations. Each scene contains 6 images, related by ground truth homographies.

\subsubsection{Evaluation Protocol.}
We follow prior work \cite{LoFTR, SiLK} and resize the shorter image side to 480 pixels. We report the repeatability metric as in \cite{Superpoint, SiLK} to evaluate interest point detection quality. We assess keypoint descriptors with the mean matching accuracy (MMA) - the ratio of matches with reprojection error below a threshold \cite{DISK}, and the matching score - the ratio of correctly matched points to total detected points. We use OpenCV's RANSAC to estimate the homography between image pairs and report the homography estimation accuracy and area under the curve (AUC) error over all image pairs. We retain the top 1k detected points for SuperPoint and 10k for SiLK.

\subsubsection{Baselines.}

We assess SuperPoint-PrP by comparing it to the baseline SuperPoint model \cite{Superpoint}. We also introduce a hybrid variant, "SP-PrP-Hyb," where the MagicPoint model undergoes training on two rounds of Homographic Adaptation before training the SuperPoint model on the NeRF dataset using the PrP process and Projective Adaptation. To assess the impact of the dataset size, we include two additional variants: "SP-(10k COCO)" a version of the baseline model trained on 10k MS-COCO images matching the size of the NeRF dataset, and "SP-PrP-(2k)" a version of SuperPoint-PrP trained on 2k NeRF images (200 randomly sampled images from each scene) instead of the full NeRF dataset. Finally, we present two PrP SiLK models: "SiLK-PrP," trained without rotation or scaling augmentation as detailed in Sec. \ref{Implementation-details}, and "SiLK-PrP-Aug," which includes the aforementioned augmentations during training.

\subsubsection{Results.}
\begin{figure}[tb]
  \centering
  \begin{subfigure}{0.3\linewidth}
    \includegraphics[height=7.0cm]{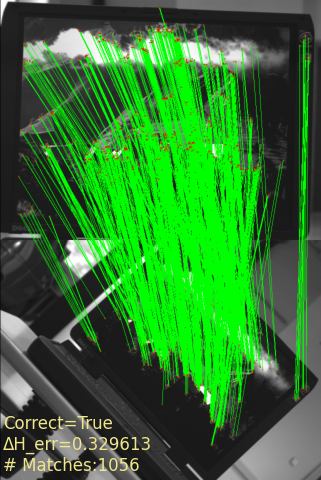}
    \caption{SiLK \cite{SiLK}}
    \label{fig:H-a}
  \end{subfigure}
  \begin{subfigure}{0.3\linewidth}
    \includegraphics[height=7.0cm]{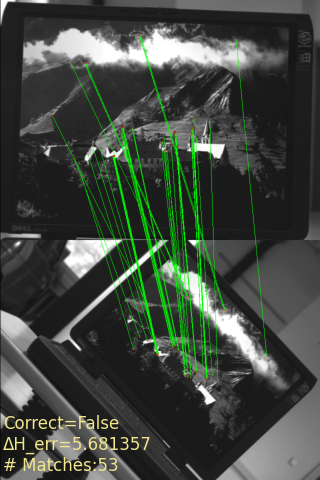}
    \caption{SiLK-PrP}
    \label{fig:H-b}
  \end{subfigure}
  \begin{subfigure}{0.3\linewidth}
    \includegraphics[height=7.0cm]{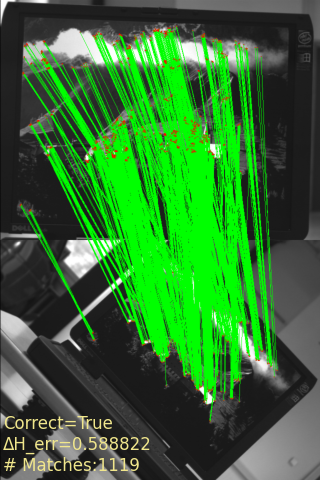}
    \caption{SiLK-PrP-Aug}
    \label{fig:H-c}
  \end{subfigure}
  \caption{\textbf{Rotation and Scale invariance.} In Fig. \ref{fig:H-b}, it is evident that SiLK-PrP lacks rotation and scale invariance compared to SiLK (\textit{cf.} Fig. \ref{fig:H-a}) \cite{SiLK}. Incorporating rotation and scaling augmentations during SiLK-PrP's training, rotation and scale invariance is achieved (\textit{cf.} Fig \ref{fig:H-c}).}
  \label{fig:Homography}
\end{figure}

As seen in Tab. \ref{tab:HPatches}, the PrP-supervised models underperform compared to their baselines \cite{SiLK, Superpoint} across all HPatches metrics. This is expected as both SuperPoint-PrP and SiLK-PrP are not rotation or scale-invariant. Despite SiLK-PrP-Aug's efforts to incorporate in-place augmentation for rotation and scale invariance (\textit{cf.} Fig. \ref{fig:Homography}), it still lags behind SiLK \cite{SiLK}. The SP-(10k COCO) model shows a significant performance drop compared to the baseline SuperPoint \cite{Superpoint}, while SP-PrP-(2k) remains competitive with SP-PrP, highlighting the advantages of the multi-view NeRF dataset and the robustness of the PrP supervision.

Nonetheless, Tab. \ref{tab:Hpatches_separated} reveals that SiLK-PrP-Aug's homography estimation is consistent with SiLK \cite{SiLK} in varying viewpoints but drops in scenes with varying illumination. This discrepancy contributes to the observed performance gap between SiLK-PrP-Aug and SiLK \cite{SiLK} in Tab. \ref{tab:HPatches}. Although the same photometric augmentation as SiLK \cite{SiLK} was used for training SiLK-PrP and SiLK-PrP-Aug, optimising these parameters during training SiLK-PrP and SiLK-PrP-Aug on the NeRF dataset may yield improved homography estimation accuracy and greater robustness to changes in illumination.

\begin{table}[tb]
    \caption{\textbf{HPatches Metrics.} Both baseline SiLK \cite{SiLK} and SuperPoint \cite{Superpoint}, surpass SiLK-PrP and SuperPoint-PrP on HPatches metrics.} 
    \label{tab:HPatches}
    \centering
    \begin{tabular}{c|cc|cc|cc|cc|c}
    \hline
    & \multicolumn{2}{c|}{Rep. $\uparrow$}       & \multicolumn{2}{c|}{Hom. Est. Acc. $\uparrow$} & \multicolumn{2}{c|}{Hom. Est. AUC $\uparrow$} & \multicolumn{2}{c|}{MMA $\uparrow$}        & MS $\uparrow$ \\ \hline
    & $\epsilon$ = 3 & $\epsilon$ = 5 & $\epsilon$ = 3   & $\epsilon$ = 5   & $\epsilon$ = 3   & $\epsilon$ = 5  & $\epsilon$ = 3 & $\epsilon$ = 5 &    \\ \hline

    SP  &   \underline{0.58}   &   0.67 &   \underline{0.81}   &   \underline{0.87}  &    \underline{0.55}     &   \underline{0.67}     &   \underline{0.68}   &  0.73  & 0.52\\

    SP-(10k COCO)  &  0.51  &  0.62  &   0.77   &   0.83  &  0.52 &  0.63  &   0.63 & 0.69   & 0.49\\
    
    SP-PrP-Hyb &  \underline{0.58}    &    \underline{0.70}   &  0.79    &    \underline{0.87}    &    0.51     &   0.64     &   \underline{0.68}   &  \underline{0.75}  &  \underline{0.55}  \\
    
    SP-PrP   &  0.53    &   0.67    &   0.75   &  0.84   &    0.45     &      0.60  &  0.65    & 0.73 &  0.51  \\
    
    SP-PrP-(2k)   &   0.52   &   0.67    &   0.75   &  0.83   &   0.44    &   0.59   &    0.64   &  0.71 &  0.49  \\\hline 
    
    SiLK   &   \underline{0.77}   &    \underline{0.85}   &  \underline{0.84}    &   \underline{0.90}     &     \underline{0.65}    &     \underline{0.74}   &   \underline{0.66}   & \underline{0.67}  &  \underline{0.38}  \\
    
    SiLK-PrP   &   0.72   &   0.80    &  0.75    &   0.81   &    0.54     &   0.64     &   0.62   & 0.63   &  0.31  \\
    
    SiLK-PrP-Aug    &   0.74   &    0.82   &  0.78    &   0.85     &    0.56     &   0.67    &  0.60    &  0.61  &  0.32  \\ \hline
    \end{tabular}
\end{table}

\begin{table}[tb]
\centering
\caption{\textbf{HPatches metrics separated.} SiLK-PrP and SiLK are similar in estimating homography for scenes with varying viewpoints, while SiLK-PrP falls behind in scenes with varying illumination.}
\label{tab:Hpatches_separated}
\begin{tabular}{c|cccc|cccc}
\hline
             & \multicolumn{4}{c|}{HPatches Viewpoint}                                                     & \multicolumn{4}{c}{HPatches Illumination}                                                  \\ \hline
             & \multicolumn{2}{c|}{Rep. $\uparrow$}                 & \multicolumn{2}{c|}{Hom. Est. Acc. $\uparrow$} & \multicolumn{2}{c|}{Rep. $\uparrow$}                 & \multicolumn{2}{c}{Hom. Est. Acc. $\uparrow$} \\ \hline
             & $\epsilon$ = 3 & \multicolumn{1}{c|}{$\epsilon$ = 5} & $\epsilon$ = 3         & $\epsilon$ = 5        & $\epsilon$ = 3 & \multicolumn{1}{c|}{$\epsilon$ = 5} & $\epsilon$ = 3        & $\epsilon$ = 5        \\ \hline
SiLK         & \underline{0.79}           & \multicolumn{1}{c|}{0.86}           & \underline{0.73}                   & \underline{0.84}                  & \underline{0.75}           & \multicolumn{1}{c|}{\underline{0.84}}           & \underline{0.95}                  & \underline{0.96}                  \\
SiLK-PrP-Aug & \underline{0.79}           & \multicolumn{1}{c|}{\underline{0.87}}           & 0.71                   & 0.81                  & 0.69           & \multicolumn{1}{c|}{0.77}           & 0.87                  & 0.89                  \\ \hline
\end{tabular}
\end{table}

\subsection{Relative Pose Estimation}\label{RelativePoseEst}

We leverage the ScanNet dataset \cite{Scannet} to evaluate the relative pose estimation for indoor scenes, employing the test split of 1500 image pairs provided by \cite{SuperGlue}. 

For outdoor relative pose estimation, the YFCC100M \cite{YFCC} and MegaDepth \cite{Megadepth} datasets are utilised. We adopt the YFCC100M test split introduced by \cite{SuperGlue}, which includes 1000 image pairs sampled from four scenes, totalling 4000 image pairs. Additionally, we use the MegaDepth1500 test split, introduced by \cite{DISK, LoFTR}, which consists of 1500 image pairs from two phototourism scenes: “Sacre Coeur” and “St. Peter’s Square”.

\subsubsection{Evaluation Protocol.}

For indoor relative pose evaluation, we follow \cite{LoFTR, SuperGlue}, resizing ScanNet images to [640 $\times$ 480]. We use the top 1k predicted points for SuperPoint and 20k for SiLK. 

For the outdoor pose evaluation, the YFCC100M and MegaDepth1500 images are resized so that their longest dimensions are 1200 and 1600 pixels, respectively. We use the top 2k points for SuperPoint and 20k for SiLK.

In both cases, we report the pose error AUC at thresholds ($5^{\circ}$, $10^{\circ}$, $20^{\circ}$), where pose error is the maximum between the angular rotation and translation errors. We also compute the essential matrix using OpenCV's RANSAC with a threshold of 0.5 over the mean focal length.

\subsubsection{Baselines.}

As in Sec. \ref{Homography_est}, we compare SuperPoint-PrP, SuperPoint-PrP-Hyb, SuperPoint-PrP-(2k) and SuperPoint-(10k COCO) with the baseline SuperPoint model, likewise, we assess SiLK-PrP and SiLK-PrP-Aug against the baseline SiLK model.

\subsubsection{Indoor Pose Estimation Results.}

As depicted in Tab. \ref{tab:IndoorPose}, the PrP models consistently outperform their baselines across all angular pose error thresholds. While SuperPoint-PrP and SuperPoint-PrP-Hyb show only slight improvement over SuperPoint \cite{Superpoint}, SiLK-PrP and SiLK-PrP-Aug display a more notable gain, especially at higher angular thresholds. Additionally, SuperPoint-PrP-(2k) and SuperPoint-(10k COCO) follow the same trend as in Sec. \ref{Homography_est}, where they underperform compared to both baseline SuperPoint and SuperPoint-PrP, with SuperPoint-(10k COCO) displaying the weakest performance.

Unfortunately, the relative pose estimation computes the translation error as the angular translation between ground truth and estimated translation vectors. However, this metric is unstable due to scale factor ambiguity \cite{Multi-View-G-CV,learning-good-corr}. As shown in Fig. \ref{fig:translation_ambiguity}, the angular translation error becomes unstable when the norm of the ground truth relative translation vector ($||t_{GT}||$) is approximately 0.2 or less.

To address this issue, we refined the indoor relative pose estimation assessment.  For scenes where $||t_{GT}||$ is below 0.15, we only report pose error AUC based on angular rotation error. For scenes with $||t_{GT}||$ above 0.15, we use the maximum of angular rotation and translation errors. As presented in Tab. \ref{tab:IndoorPose_Sep}, all models excel at estimating the angular rotation in scenes with minimal translation between viewpoints, with the PrP models slightly outperforming their baselines. In scenes where $||t_{GT}||$ exceeds the defined threshold, the PrP models show further improvement compared to results reported in Tab. \ref{tab:IndoorPose}, while SuperPoint-PrP-(2k) and SuperPoint-(10k COCO) continue to lag, with the latter performing the worst.

\begin{table}[tb]
    \begin{minipage}{0.5\linewidth}
            \centering
            \includegraphics[height=5cm]{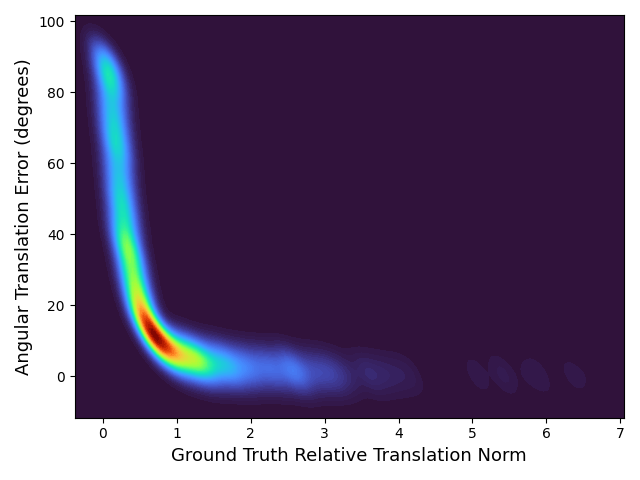}
            \captionof{figure}{\textbf{Angular Translation Error Instability.} Density plot illustrating how the angular translation error is unstable in situations when the ground truth relative translation between two camera viewpoints is minimal.}
            \label{fig:translation_ambiguity}
        \end{minipage}%
    \hfill
    \begin{minipage}{.5\linewidth}
        \centering
        \captionof{table}{\textbf{Indoor Relative Pose Estimation Error AUC.} The models trained using the PrP process outperform their respective baseline models \cite{SiLK,Superpoint} in accurately estimating relative poses within indoor scenes for all reported angular thresholds.}
        \label{tab:IndoorPose}
        \begin{tabular}{c|c|c|l}
        \hline 
        \multicolumn{1}{r|}{Pose Est. AUC} & \multicolumn{1}{l|}{@$5^{\circ}$ $\uparrow$} & \multicolumn{1}{l|}{@$10^{\circ}$ $\uparrow$} & @$20^{\circ}$ $\uparrow$ \\ \hline
        SP & 7.56  & 17.76 & 30.43 \\
        SP-(10k COCO) & 6.57 & 15.55 & 27.30  \\
        SP-PrP-Hyb  & \underline{7.82} & \underline{17.85}  &  \underline{31.08} \\
        SP-PrP    & 7.73  & 17.72   &  30.71  \\
        SP-PrP-(2k)   &  6.88 &  15.90  &  28.06 \\ \hline
        SiLK & 7.18 & 16.14 & 27.84  \\
        SiLK-PrP & 9.42 & 20.17 & 32.31 \\
        SiLK-PrP-Aug & \underline{9.60} & \underline{20.75} & \underline{34.24} \\ \hline
        \end{tabular}
    \end{minipage}
    
\end{table}

\begin{table}[tb]
\centering
\caption{\textbf{Separated Indoor Relative Pose Estimation Error AUC.} For image pairs with minimal relative translation in viewpoints, the PrP models exhibit superior performance in accurately estimating the relative rotation, especially SuperPoint-PrP and SiLK-PrP-Aug.}
\label{tab:IndoorPose_Sep}
\centering
\begin{tabular}{c|ccc|ccc}
\hline
Pose Est. AUC & \multicolumn{3}{c|}{\begin{tabular}[c]{@{}c@{}}Pose Est. AUC $||t_{gt}|| \leq \epsilon$\\ Rot.,\\ 157 image pairs\end{tabular}}      & \multicolumn{3}{c}{\begin{tabular}[c]{@{}c@{}}Pose Est. AUC $||t_{gt}|| > \epsilon$\\ $\max{(\text{Rot.},\text{Transl.})}$, \\ 1363 image pairs\end{tabular}} \\ \hline
                     & \multicolumn{1}{c|}{@$5^{\circ} \uparrow$} & \multicolumn{1}{c|}{@$10^{\circ} \uparrow$} & @$20^{\circ} \uparrow$ & \multicolumn{1}{c|}{@$5^{\circ} \uparrow$}        & \multicolumn{1}{c|}{@$10^{\circ} \uparrow$}       & @$20^{\circ} \uparrow$       \\ \hline
SP & \multicolumn{1}{c|}{55.29} & \multicolumn{1}{c|}{73.85} &84.81  &
     \multicolumn{1}{c|}{8.08}  & \multicolumn{1}{c|}{18.79} & 31.74  \\

SP-(10k COCO) & \multicolumn{1}{c|}{50.06} & \multicolumn{1}{c|}{69.14} & 82.47 & \multicolumn{1}{c|}{6.93}  & \multicolumn{1}{c|}{16.42} &  28.40 \\

SP-PrP-Hyb & \multicolumn{1}{c|}{\underline{56.17}} & \multicolumn{1}{c|}{\underline{74.21}} &\underline{85.85} & 
             \multicolumn{1}{c|}{\underline{8.37}} & \multicolumn{1}{c|}{\underline{18.83}} & \underline{32.02} \\

SP-PrP   & \multicolumn{1}{c|}{53.21} & \multicolumn{1}{c|}{73.52} &  85.28  & 
            \multicolumn{1}{c|}{8.29} & \multicolumn{1}{c|}{18.21} &  30.40  \\ 
SP-PrP-(2k)   & \multicolumn{1}{c|}{51.89} & \multicolumn{1}{c|}{72.86} &  85.17  & \multicolumn{1}{c|}{7.33} & \multicolumn{1}{c|}{16.78} &   28.93 \\ \hline

SiLK & \multicolumn{1}{c|}{54.33}  & \multicolumn{1}{c|}{73.02}   &  85.24 & \multicolumn{1}{c|}{7.62} & \multicolumn{1}{c|}{17.02} & 28.82\\
SiLK-PrP  & \multicolumn{1}{c|}{57.59}  & \multicolumn{1}{c|}{74.19}   &        85.28      & \multicolumn{1}{c|}{\underline{10.22}}  & \multicolumn{1}{c|}{21.53} & 33.52 \\
SiLK-PrP-Aug & \multicolumn{1}{c|}{\underline{57.60}} & \multicolumn{1}{c|}{\underline{75.54}} &  \underline{86.38} & \multicolumn{1}{c|}{\underline{10.22}} & \multicolumn{1}{c|}{\underline{21.79}}  & \underline{35.07} \\ \hline
\end{tabular}
\end{table}

\subsubsection{Outdoor Pose Estimation Results.}

As seen in Tabs. \ref{tab:OutdoorPose} and \ref{tab:OutdoorPose2}, the PrP models outperform their baseline counterparts in outdoor relative pose estimation, mirroring the indoor results. While, SuperPoint-PrP-(2k) and SuperPoint-(10k COCO) continue to underperform compared to other models. The results reported in Tabs. \ref{tab:IndoorPose} to \ref {tab:OutdoorPose2} consistently highlight that the PrP-supervised models show significant improvement in relative pose estimation capabilities.

\begin{table}[tb]
    \begin{minipage}{.45\linewidth}
        \centering
        \captionof{table}{\textbf{YFCC Relative Pose \\Estimation Error AUC.}}
        \label{tab:OutdoorPose}
        \begin{tabular}{c|c|c|l}
        \hline
        \multicolumn{1}{r|}{Pose Est. AUC} & \multicolumn{1}{l|}{@$5^{\circ}$ $\uparrow$} & \multicolumn{1}{l|}{@$10^{\circ}$ $\uparrow$} & @$20^{\circ}$ $\uparrow$ \\ \hline
        SP & 12.01 & 22.66 & 35.48 \\
        SP-(10k COCO) & 9.85 & 19.63 & 32.03 \\
        SP-PrP-Hyb & \underline{14.04} & \underline{26.53} & \underline{40.80} \\
        SP-PrP & 12.57 & 23.86 & 37.88 \\ 
        SP-PrP-(2k) & 11.26 & 22.72 & 36.78 \\ \hline
        SiLK & 7.82 & 15.05 & 24.93 \\
        SiLK-PrP & \underline{9.44} & \underline{17.31} & \underline{27.04} \\
        SiLK-PrP-Aug & 7.82 & 14.33 & 22.52 \\ \hline
        \end{tabular}
    \end{minipage}%
    \hfill
    \begin{minipage}{.45\linewidth}
        \centering
        \captionof{table}{\textbf{MegaDepth Relative Pose \\Estimation Error AUC.}}
        \label{tab:OutdoorPose2}
        \begin{tabular}{c|c|c|l}
        \hline
        \multicolumn{1}{r|}{Pose Est. AUC} & \multicolumn{1}{l|}{@$5^{\circ}$ $\uparrow$} & \multicolumn{1}{l|}{@$10^{\circ}$ $\uparrow$} & @$20^{\circ}$ $\uparrow$ \\ \hline
        SP & 25.21 & 39.42 & 52.11\\
        SP-(10k COCO) & 22.02 & 35.62    & 48.71\\
        SP-PrP-Hyb & \underline{27.34} & \underline{42.64} & \underline{56.03} \\
        SP-PrP & 26.17 & 41.53 & 55.87 \\
        SP-PrP-(2k) & 22.44 & 36.82   &  50.27 \\ \hline
        SiLK & 22.70 & 35.32 & \underline{48.08} \\
        SiLK-PrP & \underline{25.91} & \underline{35.50} & 46.41 \\
        SiLK-PrP-Aug & 24.79 & 34.38 & 45.00 \\ \hline
        \end{tabular}
    \end{minipage}
\end{table}

\subsection{Pairwise Point Cloud Registration}
We follow recent work \cite{unsupervisedrr, SiLK} and evaluate the interest point detectors and descriptors on a 3D point cloud registration task using ScanNet's \cite{Scannet} official test split. For the evaluation, image pairs are sampled 20 frames apart \cite{unsupervisedrr}.

\subsubsection{Evaluation Protocol.}

For point cloud registration evaluation, ScanNet \cite{Scannet} images are resized to [128 $\times$ 128]. A 6-DOF pose is estimated to align pairs of RGB-D images. We report angular rotation error (in degrees), translation error (in centimetres), and chamfer distance between ground truth and reconstructed point clouds (in centimetres). The mean and median errors are reported across the entire dataset, as well as the accuracy at different thresholds \cite{unsupervisedrr}.
 
\subsubsection{Baselines.}
Unlike previous evaluations (Secs. \ref{Homography_est} and \ref{RelativePoseEst}) where sparse feature matching was employed, we follow \cite{unsupervisedrr, SiLK} and use dense feature matching for the point cloud registration evaluation. The ratio test is utilised to find correspondences, while alignment is seen as a Procustes problem solved by a weighted Kabsch's algorithm. \cite{Procustes,global-registration, SiLK}.

\subsubsection{Results.}

As depicted in Tab. \ref{tab:pointcloud}, both SuperPoint-PrP and SuperPoint-PrP-Hyb outperform the baseline SuperPoint model across all metrics, except for translation at a 5 cm threshold. Models trained on smaller datasets slightly underperform, consistent with previous evaluations. SiLK maintains a 1-3$\%$ edge over SiLK-PrP and SiLK-PrP-Aug, with a 5$\%$ lead at the 5 cm translation threshold, though the SiLK-PrP models remain competitive.

It is essential to highlight that SiLK is trained on smaller images [164x164], while the NeRF dataset is comprised of larger images [640x480]. Since the point cloud registration is conducted on an image size of [128x128], SiLK may outperform SiLK-PrP at lower resolutions, however, the SiLK-PrP models still detect high-quality interest points at low resolutions.

\begin{table}[tb]
    \caption{\textbf{Point Cloud Reg.} SuperPoint-PrP and SuperPoint-PrP demonstrate marginal enhancements over the baseline SuperPoint model. Meanwhile, SiLK-PrP and SiLK-PrP-Aug maintain competitiveness with the baseline SiLK model.}
    \label{tab:pointcloud}
    \centering
    \scalebox{0.91}{
    \begin{tabular}{c|ccccc|cccll|lllll}
    \hline
      & \multicolumn{5}{c|}{Rot.}                                    & \multicolumn{5}{c|}{Transl.}                                                            & \multicolumn{5}{c}{Chamfer}                                                                                                  \\ \hline
      & \multicolumn{3}{c|}{Acc. $\uparrow$}        & \multicolumn{2}{c|}{Err. $\downarrow$} & \multicolumn{3}{c|}{Acc. $\uparrow$}        & \multicolumn{2}{c|}{Err. $\downarrow$}                            & \multicolumn{3}{c|}{Acc. $\uparrow$}                                               & \multicolumn{2}{c}{Err. $\downarrow$}                            \\ \hline
      & $5^{\circ}$ & $10^{\circ}$ & \multicolumn{1}{c|}{$45^{\circ}$} & M.        & Med.        & 5 & 10 & \multicolumn{1}{c|}{25} & \multicolumn{1}{c}{M.} & \multicolumn{1}{c|}{Med.} & \multicolumn{1}{c}{1} & \multicolumn{1}{c}{5} & \multicolumn{1}{c|}{10} & \multicolumn{1}{c}{M.} & \multicolumn{1}{c}{Med.} \\ \hline 
      
    SP & 88.4 & 95.1 & \multicolumn{1}{c|}{98.8} & 4.0 & \underline{1.8}
       & \underline{53.2} & \underline{79.9} & \multicolumn{1}{c|}{94.5} & 9.5 & \underline{4.7}
       & \underline{73.2} & 91.6 & \multicolumn{1}{l|}{94.8} & 6.4  & \underline{   0.4}  \\

    SP-(10k C.) & 86.1 & 93.6 & \multicolumn{1}{c|}{98.6} & 4.6 & 1.9
       & 50.7 & 77.4 & \multicolumn{1}{c|}{93.2} & 10.7 & \underline{4.9}
       & 70.6 & 90.0 & \multicolumn{1}{l|}{93.6} &  7.3 &    \underline{  0.4} \\
    
    SP-PrP-Hyb & \underline{88.5} & \underline{95.9} & \multicolumn{1}{c|}{\underline{99.1}} & \underline{3.8} & 1.9 
             & 50.9 & 79.5 & \multicolumn{1}{c|}{\underline{95.5}} & \underline{8.9} & 4.9 
             & 71.9 & \underline{92.4} & \multicolumn{1}{l|}{\underline{95.6}} & \underline{5.6} & \underline{  0.4} \\
             
    SP-PrP   & 88.3 & \underline{95.9} & \multicolumn{1}{c|}{\underline{99.1}} & \underline{3.8} & 1.9
             & 51.4 & \underline{79.9} & \multicolumn{1}{c|}{95.3} & 9.0 & 4.9
             & 72.4 & 92.2 & \multicolumn{1}{l|}{95.5} & \underline{5.6} & \underline{  0.4} \\
    
    SP-PrP-(2k)   & 87.9 & 95.5 & \multicolumn{1}{c|}{99.0} & 4.0 & 1.9
             & 50.9 &  79.2 & \multicolumn{1}{c|}{95.1} & 9.4 & 4.9
             &  72.0 & 92.1 & \multicolumn{1}{l|}{95.4} & \underline{5.7} & \underline{  0.4} \\ \hline
           
    SiLK  & \underline{95.6} & \underline{97.5} & \multicolumn{1}{c|}{\underline{99.1}} & \underline{2.6} & \underline{0.8}      
          & \underline{79.9} &  \underline{91.4} & \multicolumn{1}{c|}{\underline{96.9}} & \underline{6.1} & \underline{2.2}
          & \underline{89.4} & \underline{95.8} & \multicolumn{1}{l|}{\underline{97.1}} & \underline{4.7} & \underline{   0.1}  \\
          
    SiLK-PrP & 93.0 & 96.2 & \multicolumn{1}{c|}{98.7} & 3.4 & 1.0 
             & 71.3 & 87.0 & \multicolumn{1}{c|}{95.3} & 8.1 & 2.8 
             & 84.3 & 93.7 & \multicolumn{1}{l|}{95.7} & 5.7 & \underline{   0.1}  \\
    
    SiLK-PrP-A. & 94.6 & 97.2 & \multicolumn{1}{c|}{99.0} & 2.8 & 0.9                  & 75.0 & 89.5 & \multicolumn{1}{c|}{96.4} & 7.0 & 2.5                  & 86.8 & 95.1 & \multicolumn{1}{l|}{96.7} & 5.4 & \underline{   0.1}               \\ \hline
    \end{tabular}}
\end{table}

\section{Conclusion}
\label{sec:conclusion}
This paper presented a novel approach to supervise feature point detectors and descriptors using perspective projective geometry on synthetic data. Despite using a smaller dataset, our PrP approach achieves similar or better performance without compromising generalisability. Moreover, we noticed that the PrP-supervised models converge faster and outperform homography-supervised models in non-planar multi-view benchmarks, though, they slightly lag in homography estimation. We emphasise the main advantage of the PrP approach is not only benchmark improvements, but also its ability to train on any custom multi-view synthetic dataset with less data and training time. Additionally, while the real images used to render NeRF scenes could potentially train the detectors directly with COLMAP-produced data, this approach is limited. The NeRF dataset uses an average of 80-100 images per scene, totalling less than 1000 images with limited viewpoint diversity. The degradation in performance observed from SuperPoint-PrP-(2K) compared to SuperPoint-PrP suggests that training on even fewer real images would further impair model performance. The larger potential for further advancements depends on enhancing neural rendering to produce higher-quality synthetic images and more accurate depth maps, reducing artefacts and misprojection errors.

\section*{Acknowledgements}
This work was partly conducted during an MSc in Robotics and Computation at the Department of Computer Science, University College London (UCL).

\newpage
\bibliographystyle{unsrt}  
\bibliography{references}

\end{document}